\documentclass[lettersize,journal]{IEEEtran}
\usepackage{graphicx} % Required for inserting images
\usepackage{xcolor}

\DeclareMathAlphabet{\pazocal}{OMS}{zplm}{m}{n}

\usepackage{amsmath, amssymb}
\usepackage{graphicx}
\usepackage[colorlinks=false, allcolors=blue]{hyperref}
\usepackage{cite}
\usepackage{makecell}
\usepackage{booktabs}
\usepackage{enumitem}
\usepackage{caption}

\usepackage{textcomp}
\usepackage{stfloats}

\usepackage{parskip}
\usepackage{subcaption}

\begin{document}

\title{Concept backpropagation: An Explainable AI approach for visualising learned concepts in neural network models}

\author{\IEEEauthorblockN{Patrik Hammersborg\IEEEauthorrefmark{1},
Inga Strümke\IEEEauthorrefmark{2}}
\IEEEauthorblockA{\\
\IEEEauthorrefmark{1}patrik.hammersborg@ntnu.no,
\IEEEauthorrefmark{2}inga.strumke@ntnu.no}
        % <-this % stops a space
\thanks{P. Hammersborg and I. Strümke are affiliated with the Department of Computer Science,
Norwegian University of Science and Technology,
Trondheim, Norway}% <-this % stops a space
% \thanks{Manuscript received xx xx, xx; revised xx xx, xx.}
}
% The paper headers
% \markboth{IEEE Transactions on Neural Networks and Learning,~Vol.~xx, No.~xx, Month~2023}%
% {Hammersborg and Strümke: Concept backpropagation: An Explainable AI approach for visualising learned concepts in neural network models}
% \IEEEpubid{0000--0000/00\$00.00~\copyright~2023 IEEE}
% \IEEEpubid{~\copyright~2023 IEEE}
\IEEEpubid{\begin{minipage}{\textwidth}\ \\[12pt] \centering
\copyright 2023 IEEE. Personal use of this material is permitted. Permission
from IEEE must be obtained for all other uses, in any current or future
media, including reprinting/republishing this material for advertising or
promotional purposes, creating new collective works, for resale or
redistribution to servers or lists, or reuse of any copyrighted
component of this work in other works.
\end{minipage}} 
\maketitle

\begin{abstract}
Neural network models are widely used in a variety of domains, often as black-box solutions, since they are not directly interpretable for humans. The field of explainable artificial intelligence aims at developing explanation methods to address this challenge, and several approaches have been developed over the recent years, including methods for investigating what type of knowledge these models internalise during the training process. Among these, the method of concept detection \cite{tcav_2018}, investigates which \emph{concepts} neural network models learn to represent in order to complete their tasks. In this work, we present an extension to the method of concept detection, named \emph{concept backpropagation}, which provides a way of analysing how the information representing a given concept is internalised in a given neural network model. In this approach, the model input is perturbed in a manner guided by a trained concept probe for the described model, such that the concept of interest is maximised. This allows for the visualisation of the detected concept directly in the input space of the model, which in turn makes it possible to see what information the model depends on for representing the described concept.
We present results for this method applied to a various set of input modalities, and discuss how our proposed method can be used to visualise what information trained concept probes use, and the degree as to which the representation of the probed concept is entangled within the neural network model itself.
\end{abstract}


\section{Introduction}
Neural network models are becoming increasingly common for solving many complex problems. However, these models are not interpretable to humans, meaning that it is not directly knowable what information the models use to make their predictions. During recent years, methods have been developed that allow for the probing of what such models have learned, through the representation of information as \emph{concepts}. The method of concept detection is fundamentally based on ``interpreting the intermediate states of neural network in terms of human-friendly concepts" \cite{tcav_2018}. In this case, a concept is an human-defined abstraction of information present in a given input sample.\footnote{The information required to make up a concept might not be directly represented by a single input feature, but rather as a function of a set number of features. A widely used example, as presented in \cite{tcav_2018}, is the notion of ``stripes" as a concept for detecting zebras in an image.} of a  is  While concept detection is useful for probing the presence of a predefined knowledge, i.e.\ concepts, it does not provide a means for detecting exactly \emph{how} said knowledge is internalised in the model. It is possible to know whether knowledge is represented in the model, but this does not guarantee that its representation is not, e.g., entangled with some other information. 

In order to investigate how knowledge is represented in a neural network model, we propose the method of \emph{concept backpropagation}. This method allows for a visualisation of how a given concept is internalised by a neural network model. This is done by arranging a structure that allows for the maximisation of a pre-trained concept probe, i.e. being able to transform an input sample in order to maximise a given concept. This provides a means of investigating what information is being used to detect the described concept in the model, in addition to making it possible to visualise the internalisation of the concept directly in the model's input space.
We apply the method to a various set of problem cases, including tabular data, images, and chess.

The paper is structured as follows:
In Sec.~\ref{section:method}, we provide the necessary background and describe the proposed method of concept backpropagation.
In Sec.~\ref{sec:applications} we present its application on four different cases featuring different data input spaces.
In Sec.~\ref{section:results}, we show the results of the method on the presented problem cases.
In Sec.~\ref{section:discussion}, we discuss the benefits and limitations of the proposed method, in addition to highlighting how it is relevant in the broader field of Explainable Artificial Intelligence (XAI).

We also provide an open source repository containing code implementing all our described methods.\footnote{This is available at \url{https://github.com/patrik-ha/concept-backpropagation}.}

\section{Method}\label{section:method}
\subsection{Concept detection}
\IEEEpubidadjcol
Our proposed method is based on the established concept detection method used in \cite{alphazero_concepts_2021}, itself based on the work presented in \cite{tcav_2018}. In a nutshell, for a neural network model $M: I \rightarrow O$ with an intermediary layer $L$, and a concept function $f(s)$ that quantifies the presence of some concept $C$ in an input sample $s$, concept detection aims to indicate if $M$ learns to distill information pertaining to $C$ by looking at the information generated in $L(s)$, as discussed in \cite{alain2018understanding}. This is done by training a logistic probe $P: L \rightarrow C$ on a large set of samples $(L(s), f(s))$, and if successful, means that enough information pertaining to $C$ is linearly represented in the values generated in $L(s)$. 

For a binary concept, the probe is trained by minimising
\begin{equation}\label{equation:default-probe}
    \left\| {\sigma \left( {\mathbf{w} \cdot {L(s_i)} + \mathbf{b}} \right) - {f(s_i)}} \right\|_2^2 + \lambda {\left\| \mathbf{w} \right\|_1} + \lambda \left| \mathbf{b} \right|\,,
\end{equation}
for each pair $(L(s_i), f(s_i))$, where $\mathbf{w}$ and $\mathbf{b}$ are the trainable parameters of the probe, and $\sigma$ is the standard sigmoid function. The method is also adapted for scalar concepts by removing the use of the sigmoid function from Eq.~\ref{equation:default-probe}, which essentially changes the learned relationship from being logistic to being linear.

While concept detection gives a direct assessment of the presence of a described concept in a model, it does not guarantee that the information used by the concept function $f(s)$ is the same as the information being used in $L(s)$ to construct $C$. However, since these trained probes effectively indicate a learned relationship between the layer $L$ and the specific concept $C$, it is possible to use a trained probe to find an out which elements of $L$ are being used to represent $C$. Then, by using this dependency between $L$ and $C$, said information can be used to infer how changes in a given state $s$ affect the detected concept $C$. Since the probes can be represented as generic single-layer neural networks, the gradients of the probe's output can be used to guide a search for a perturbation of $s$ that maximises $f(s)$ according to $P$.

\subsection{Concept backpropagation}\label{section:concept-backpropagation}
\begin{figure}
    \centering
    \includegraphics[width=0.65\linewidth]{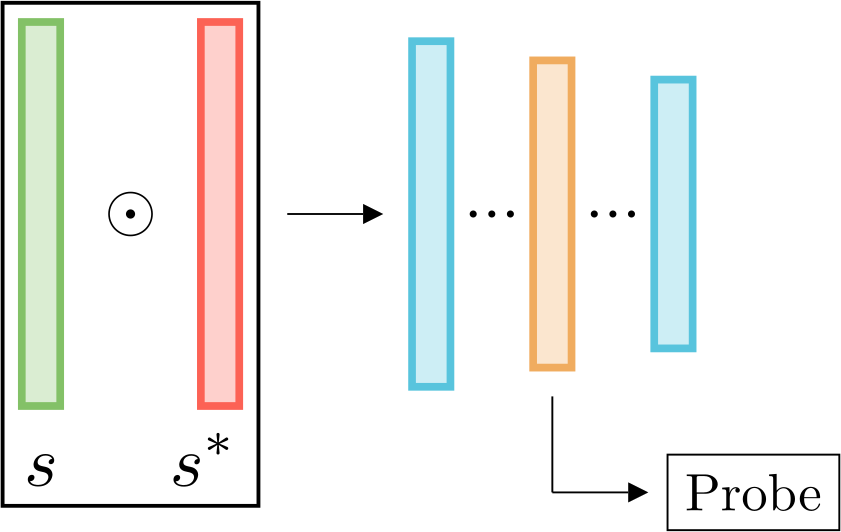}
    \caption{The architecture described in Sec.~\ref{section:concept-backpropagation}, which enables the maximisation of a given concept, by following the minimisation objective described in Eq.~\ref{equation:generic-backprop}. For a single sample $s$, the objective is to find a perturbation $s^*$ that maximises the desired probe output, where the probe provides gradients to guide the search.}
    \label{fig:main-arrangement} 
\end{figure}

Our idea is formulated as a minimisation problem. For a model $M: I \rightarrow O$, with an intermediate layer $L(s)$ and a trained logistic probe $P: L \rightarrow C$, an input state $s$, a concept output $o$ as the desired output of the concept function $f(\cdot)$, and some combination operator $\odot$, we wish to find a minimal perturbation $s^*$ so that $P(L(s \odot s^*)) = o$. That is, we aim to minimise
\begin{equation}\label{equation:generic-backprop}
    \lambda_1 |P(L(s \odot s^*)) - o| + \lambda_2 \operatorname*{dist}(s, s^*) ,
\end{equation}
where $\operatorname*{dist}(\cdot, \cdot)$ is a function that indicates the distance between $s \odot s^*$ and $s$, and $\lambda_1$, $\lambda_2$ are weighting constants, in the range $[0, \infty]$. Both $\odot$ and $\operatorname*{dist}(\cdot, \cdot)$ are chosen to suit the input space of the presented problem. A high-level illustration of the described setup is shown in Fig.~\ref{fig:main-arrangement}. This minimisation process is done by standard gradient descent, meaning that one needs to choose $\odot$ and $\operatorname*{dist}(\cdot, \cdot)$ to allow for adequate propagation of the gradient from the output of the probe.

\subsection{Use cases}\label{section:implementation}
While the method presented in Sec.~\ref{section:concept-backpropagation} is quite general, it can be demonstrated through application to specific problem cases.

\subsubsection{Tabular data}\label{section:tabular-data}
Our first use case is a neural network model $M$ trained on tabular data whose input space consists of samples of $n$-dimensional real-valued vectors. For this case, the distance function is defined as $dist(s, s^*) = ||s^*||_2$, and $\odot$ as standard, element-wise addition. This gives the following minimisation objective,
\begin{equation}\label{equation:real-valued-backprop}
    |P(L(s + s^*)) - o| + ||s^*||_2,
\end{equation}
for $\lambda_1$, $\lambda_2$ equal to $1$.
Here, for an input vector $s$, we aim to add the smallest perturbation (the notion of ``smallest" being expressed through $||s^*||_2$) to $s$ that produces the desired probe output.

\subsubsection{Images}\label{section:images}
Our next use-case are neural network models, typically convolutional neural networks (CNNs), trained to handle images. We observe that it is difficult to work with perturbations of images directly, which in turn hinders the feasibility of a direct application of the proposed method. The challenges observed occur due to the difference in dimensionality between the images and the intermediate layers, as backed up by preliminary experiments: It was observed that the size of the images provided the possibility that valid perturbations only consisted of large amounts of low magnitude noise. While these did provide valid maximisations of the probe, they do not provide relevant information regarding to how the model learned to represent the relevant concepts for standard images, i.e.\ images represented by the training data or images we can expect the model to encounter during use.

The described problem is mitigated by adding an embedding network for the given image model: For an image model $M$ and an input image $s$, one first maps $s$ to a latent space by some encoding function $s_l = E(s)$. This latent space then serves as the de facto input space for concept maximisation. Then, the image can be mapped back into its original space by some decoding function $D(s_l)$. For the proposed method, this means that we can define $\odot$ as 
\begin{equation}\label{equation:image-combination-function}
    s \odot s^* = D(E(s) + s^*),
\end{equation}
and $\operatorname*{dist}(\cdot, \cdot)$ as 
\begin{equation}\label{equation:image-distance}
    \operatorname*{dist}(s, s^*) = ||D(E(s) + s^*) - s||_2.
\end{equation}
Here, $s^*$ is an $n$-dimensional vector in the created embedding space, and the distance function expresses the squared difference between an image $s$, and the decoded image after having its embedding perturbed by $s^*$. The main idea is that the perturbation now takes place in the embedding space, circumventing the need to perturb the image in its original representation.

\subsubsection{Chess}\label{section:chess}
In our final use-case, we consider a model for playing 6x6-chess, first presented in \cite{masterpiece}. It is trained by reinforcement learning (RL) model through self-play, similar to the model being used in \cite{alphazero_concepts_2021}. In this case, many of the aspects of the method are adapted to fit the intricacies of chess as an input space. The positional aspect (i.e.~ the pieces on the board) are strictly binary, meaning that all elements of a perturbation $s^*$ need to be binary. Additionally for a state $s$, $s^*$ can only add pieces to vacant squares, or remove pieces from filled squares, which in turn places some restrictions as to what perturbations are valid for $s$. In this case, $s^*$ was decomposed into two trainable binary matrices, $s^-$ and $s^+$. $\odot$ was then defined as 
\begin{equation}\label{equation:chess-combination-function}
    s \odot s^* = (s - s^-) + s^+,
\end{equation}
 for $s^* = (s^-, s^+)$, where $s^-$ designated which squares were to have pieces removed, and $s^+$ designated which squares to have pieces added to it, in addition to what pieces should be added to the applicable square(s).\footnote{The binary nature of this mask is upheld by implementing the masks as binarised layers, as presented in \cite{binaryneuralnet}.}

Since it was desirable for $s^*$ to only give perturbations that produced legal positions within the rules of chess, the distance function was modified to accommodate this. A legality classifier $c(\cdot)$ was trained to discern legal and illegal positions of chess, and used to augment the distance estimate for $s \odot s^*$, by letting
\begin{equation}\label{equation:chess-distance-function}
    \operatorname*{dist}(s, s^*) = c(s \odot s^*) + ||s^+||_1 + ||s^-||_1.
\end{equation}
Here, the main point is that since the legality classifier itself was a neural network model, it too could produce gradients allowing its output to be minimised by finding an adequate $s^*$. Additional information regarding the implementation of $s^-$ and $s^+$, and details wrt.~ using chess as an input space can be found in \cite{mthesis}.

\section{Applications}\label{sec:applications}

\subsection{Tabular data}

\begin{table}
    \centering
    \captionsetup{width=.9\linewidth}
    \caption{\label{table:california-housing-dataset}Description of each feature in the California Housing dataset. Note that each sample does not operate on individual housing units, but rather groups of housing (referred to as ``groups"). The dataset and the corresponding feature labels were obtained through \cite{scikit-learn}.}
    \begin{tabular}{ll}
    \toprule
        Feature name & Description \\
        \midrule
        $\mathbf{MedInc}$ & \makecell[l]{Median income in group} \\[0.3cm]
        $\mathbf{HAge}$ & \makecell[l]{Median house age in group} \\[0.3cm]
        $\mathbf{AveRms}$ & \makecell[l]{Average room number per household} \\[0.3cm]
        $\mathbf{AveBedrms}$ & \makecell[l]{Average number of bedrooms for each household} \\[0.3cm]
        $\mathbf{Pop}$ &  \makecell[l]{Population of the given group} \\[0.3cm]
        $\mathbf{AveOcp}$ & Average number of household members \\[0.3cm]
        $\mathbf{Target}$ & Median house value per group \\[0.3cm]
        \bottomrule
    \end{tabular}
    
\end{table}

We apply the method as presented in Sec.~\ref{section:tabular-data} to a small neural network model trained on an altered version of the California Housing dataset, first presented in \cite{housing_dataset}. It is used as a small tabular dataset with six real-valued features, as described in Table \ref{table:california-housing-dataset}, and it was normalised to better suit regression by a neural network. We define the probed concept to be $\frac{\mathbf{AveBedrms}}{\mathbf{AveOccup}}$, i.e. a ratio proportional to the average number of bedrooms per person for each household.

\subsection{Images}

\begin{figure}
    \centering
    \includegraphics[width=0.4\textwidth]{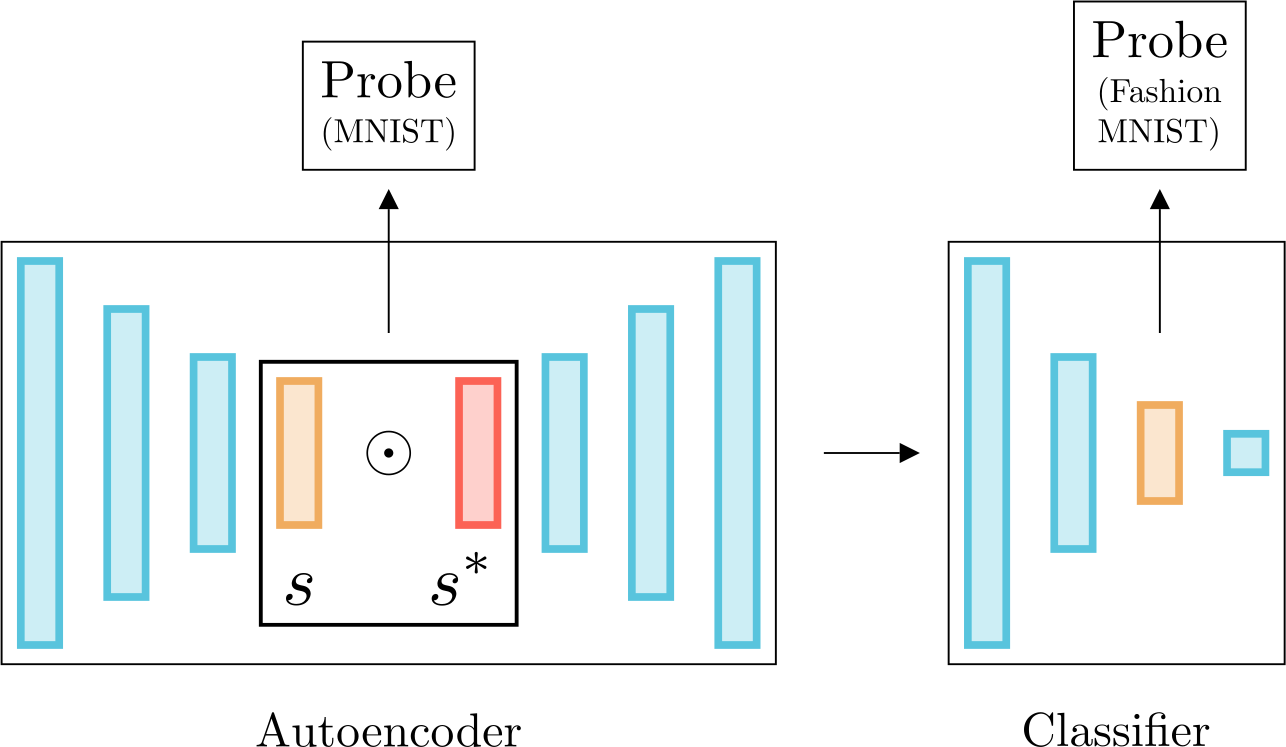}
    \caption{The arrangement of the probing and maximisation architecture for the image-based problem cases.}
    \label{fig:image-probes} 
\end{figure}

We apply the method as described in Sec.~\ref{section:images} to a convolutional autoencoder model trained on the MNIST dataset \cite{lecun2010mnist}. We aim to maximise the concept ``loopiness", i.e. if the standard drawing of the given digit includes any self-closing loops.\footnote{I.e. matching the digits $0$, $6$, $8$, $9$} We aim to maximise a concept in the latent dimension of the autoencoder itself, as shown in Fig.~\ref{fig:image-probes}.

We also apply the described method to a image classifier model trained on the Fashion-MNIST dataset. Here, the goal is to maximise the lightness of the given article of clothing, i.e. the ratio of non-black pixels to pixels with magnitude above a certain threshold. We use use a convolutional autoencoder to embed the images, and probe for the concept in an intermediate layer in the classifier, as shown in Fig.~\ref{fig:image-probes}.

\subsection{Chess}
We apply the method as presented in Sec.~\ref{section:chess} to a pre-trained model for 6x6-chess, where we seek to maximise the threat on the queen of the player to move.
\section{Results}\label{section:results}
\subsection{Tabular data}
The results for the method described in Sec.~\ref{section:tabular-data} are shown in Table \ref{table:results:housing-dataset}. In all the presented cases, we see significant changes in one or both features directly related to the concept as $\frac{\mathbf{AveBedrms}}{\mathbf{AveOcp}}$. We also observe that most of these maximisations have side-effects, changing features that are not directly correlated to this ratio, namely $\mathbf{MedInc}$ and $\mathbf{AveRms}$. This is discussed in Sec.~\ref{section:discussion}.
\begin{table}
    \centering
    \captionsetup{width=.9\linewidth}
    \caption{\label{table:results:housing-dataset}Maximisation results for five tabular samples. The results are deltas made to each input sample in order to maximise a probe trained to detect $\frac{\mathbf{AveBedrms}}{\mathbf{AveOcp}}$.}
    \scalebox{0.9}{
        \begin{tabular}{|l|l|l|l|l|l|}
        \hline
            $\mathbf{MedInc}$ & $\mathbf{HAge}$ & $\mathbf{AveRms}$ & $\mathbf{AveBrms}$ & $\mathbf{Pop}$ & $\mathbf{AveOcp}$ \\
            \hline
            $-0.009$ & $+0.000$ & $+8.999$ & $+1.560$ & $+0.121$ & $-0.223$ \\
            \hline
            $+0.000$ & $+0.003$ & $+5.067$ & $+0.811$ & $+0.126$ & $-1.216$ \\
            \hline
            $+0.000$ & $+0.057$ & $+1.797$ & $+1.803$ & $-0.567$ & $-5.696$ \\
            \hline
            $+0.013$ & $+0.003$ & $-0.003$ & $+0.000$ & $+0.003$ & $-0.798$ \\
            \hline
            $+1.125$ & $+0.011$ & $-0.007$ & $+0.001$ & $-0.061$ & $-17.494$ \\
        \hline
        \end{tabular}
    }
\end{table}

\subsection{Images}
\begin{figure}
    \centering
    \begin{subfigure}{0.3\textwidth}
        \includegraphics[width=\textwidth]{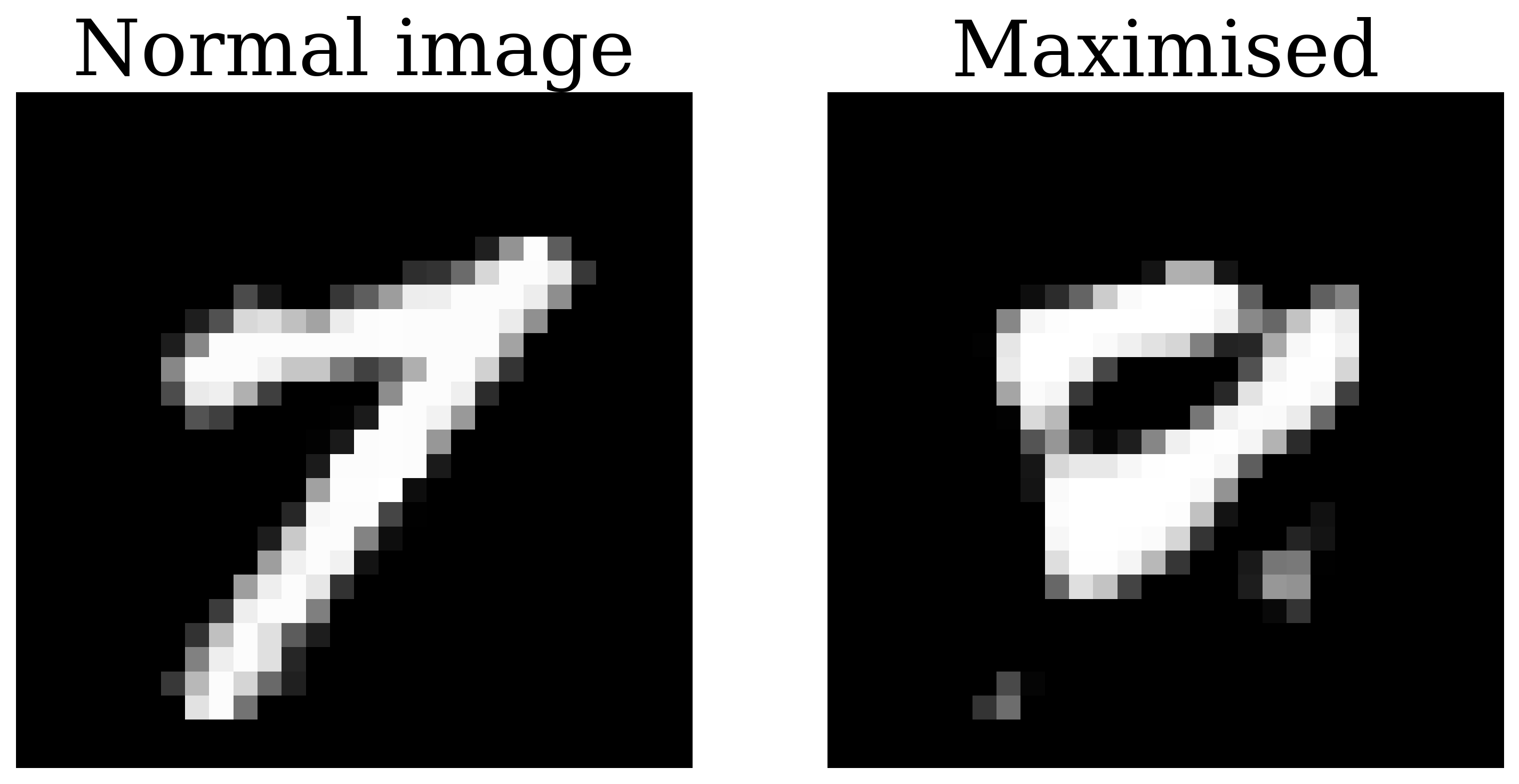}
        \caption{}
        \label{fig:mnist:0}
    \end{subfigure}
    \hspace{1.5cm}
    \begin{subfigure}{0.3\textwidth}
        \includegraphics[width=\textwidth]{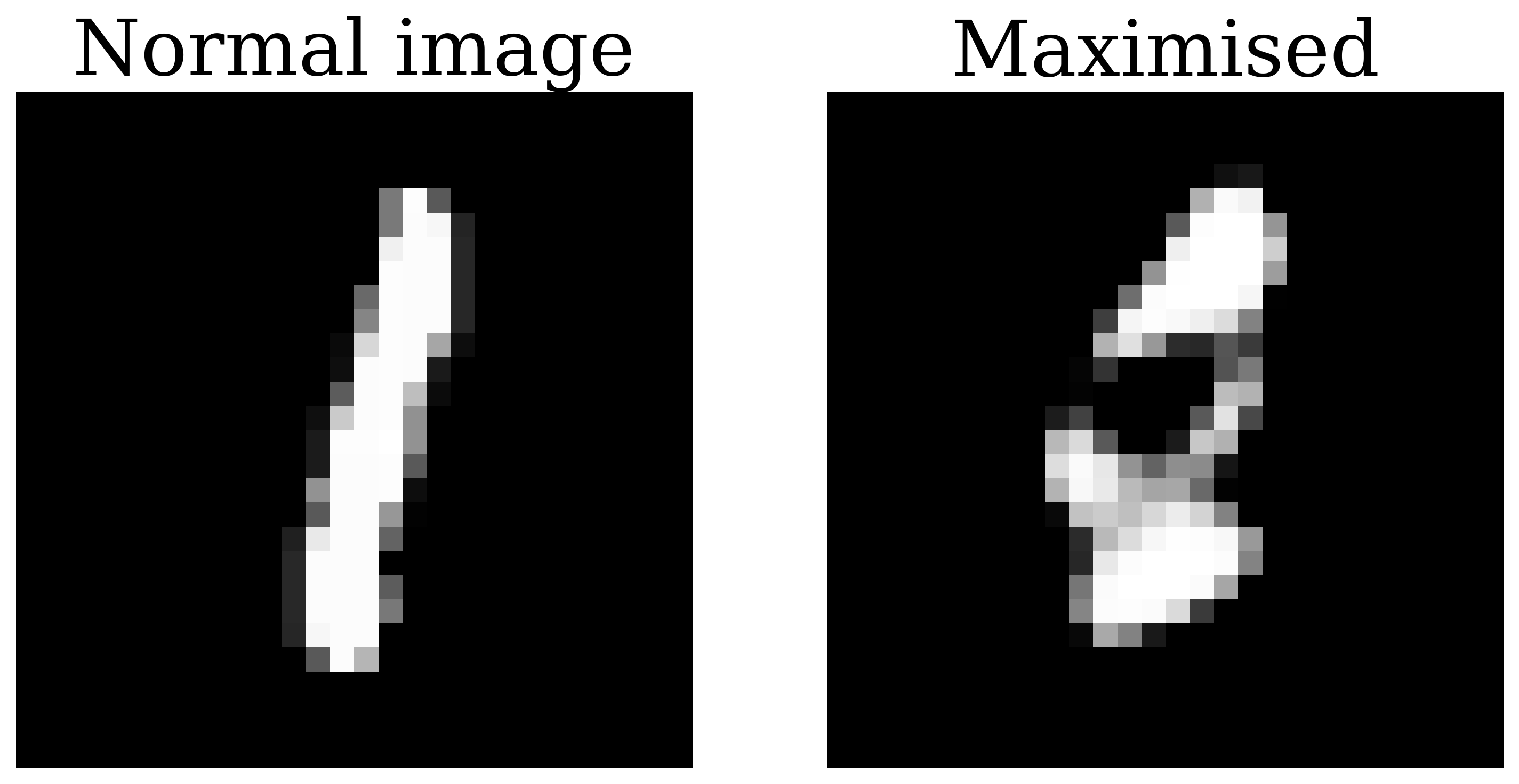}
        \caption{}
        \label{fig:mnist:1}
    \end{subfigure}
    \\[3ex]
    \begin{subfigure}{0.3\textwidth}
        \includegraphics[width=\textwidth]{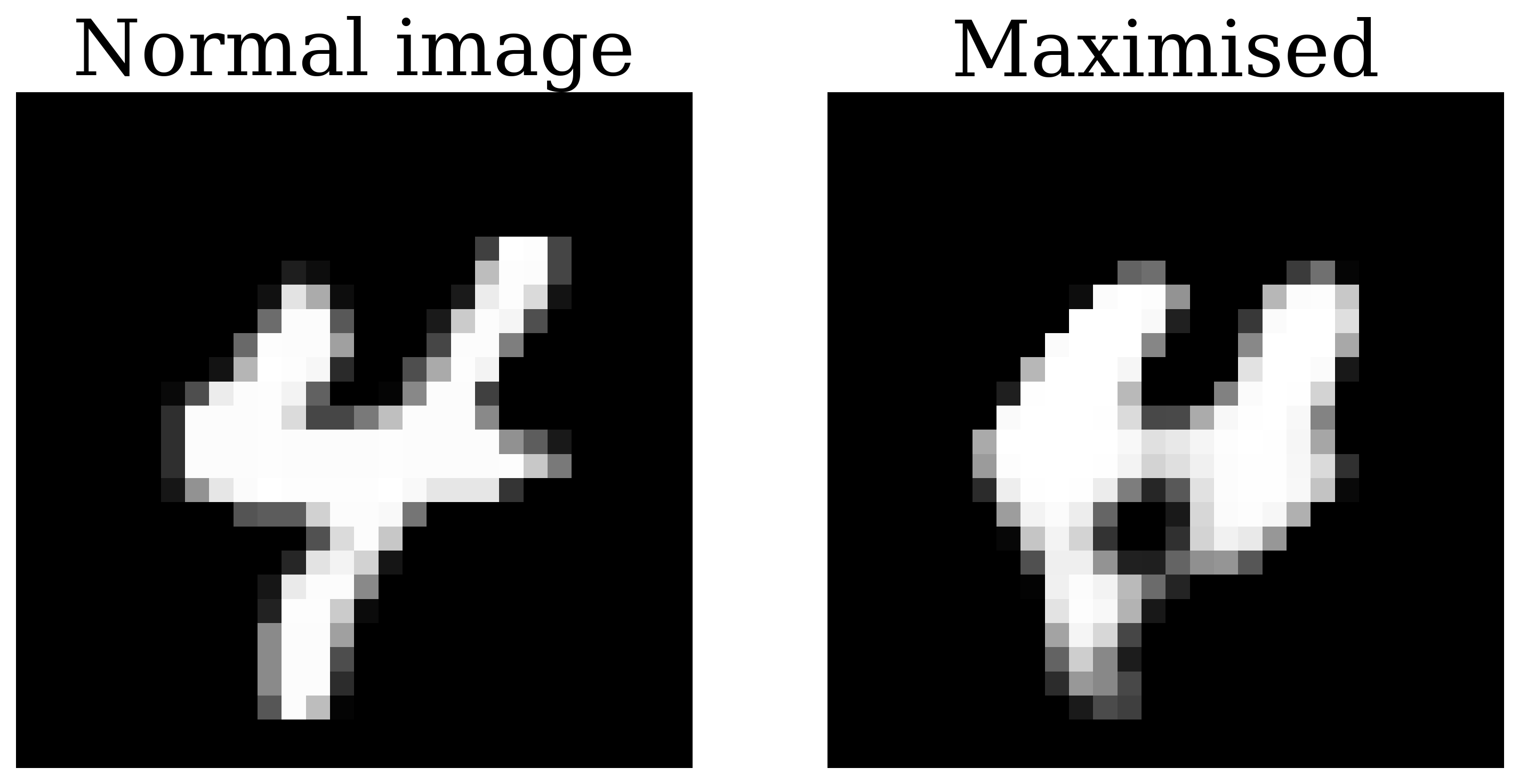}
        \caption{}
        \label{fig:mnist:2}
    \end{subfigure}
    \hspace{1.5cm}
    \begin{subfigure}{0.3\textwidth}
        \includegraphics[width=\textwidth]{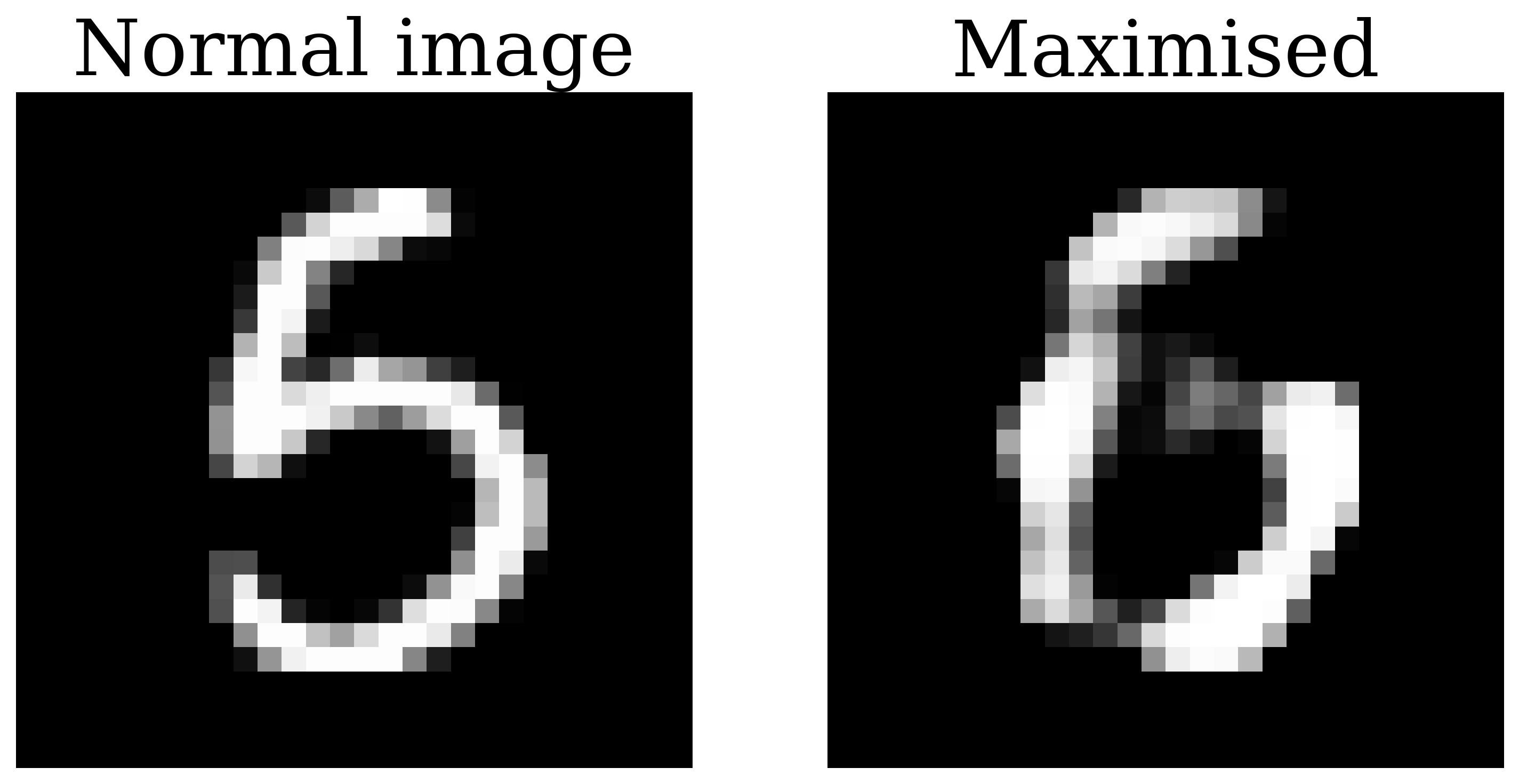}
        \caption{}
        \label{fig:mnist:3}
    \end{subfigure}
    \caption{\label{fig:mnist}Samples from the MNIST dataset, with corresponding maximisations.}
\end{figure}

\begin{figure}
    \centering
    \begin{subfigure}{0.3\textwidth}
        \includegraphics[width=\textwidth]{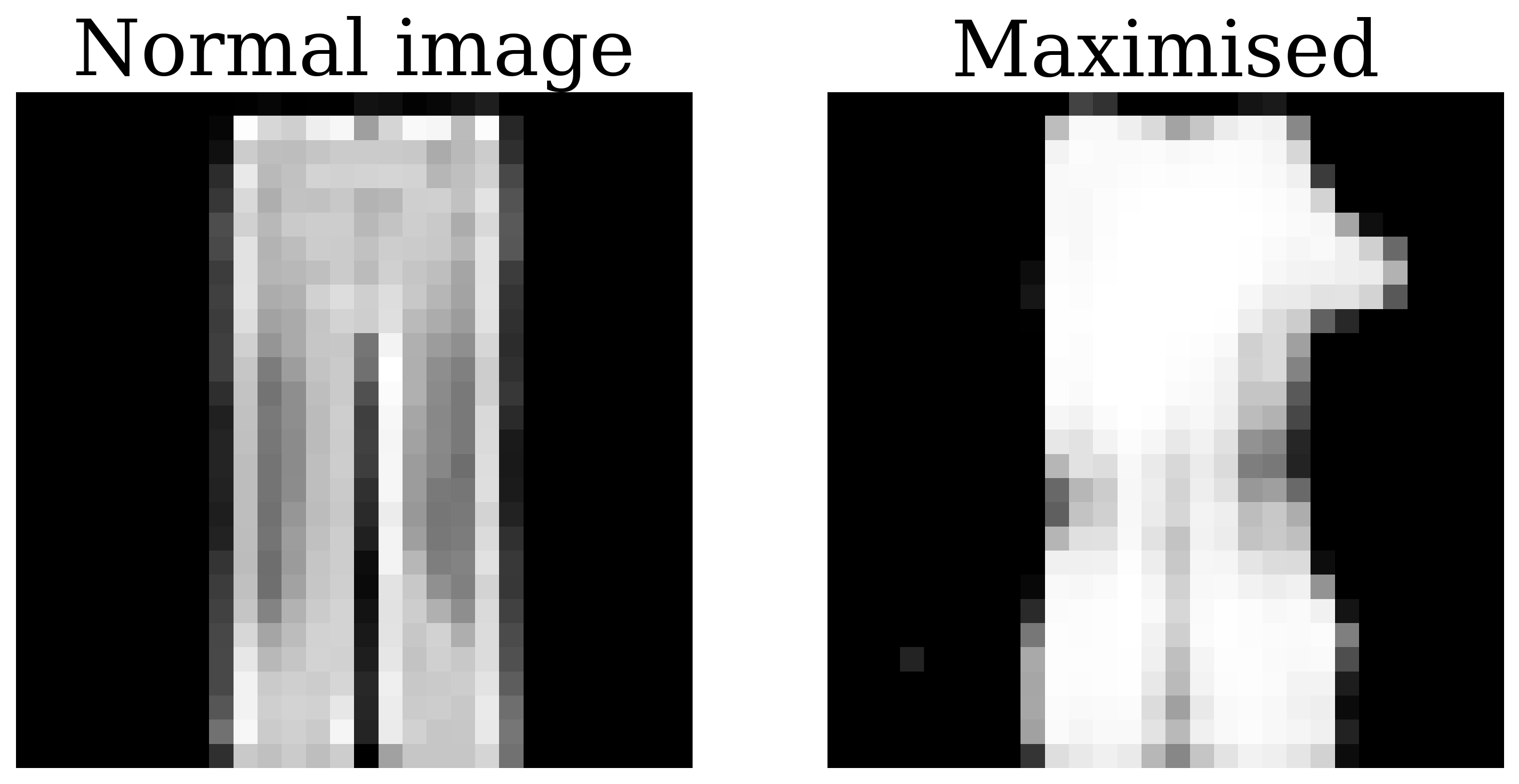}
        \caption{}
        \label{fig:fashion:0}
    \end{subfigure}
    \hspace{1.5cm}
    \begin{subfigure}{0.3\textwidth}
        \includegraphics[width=\textwidth]{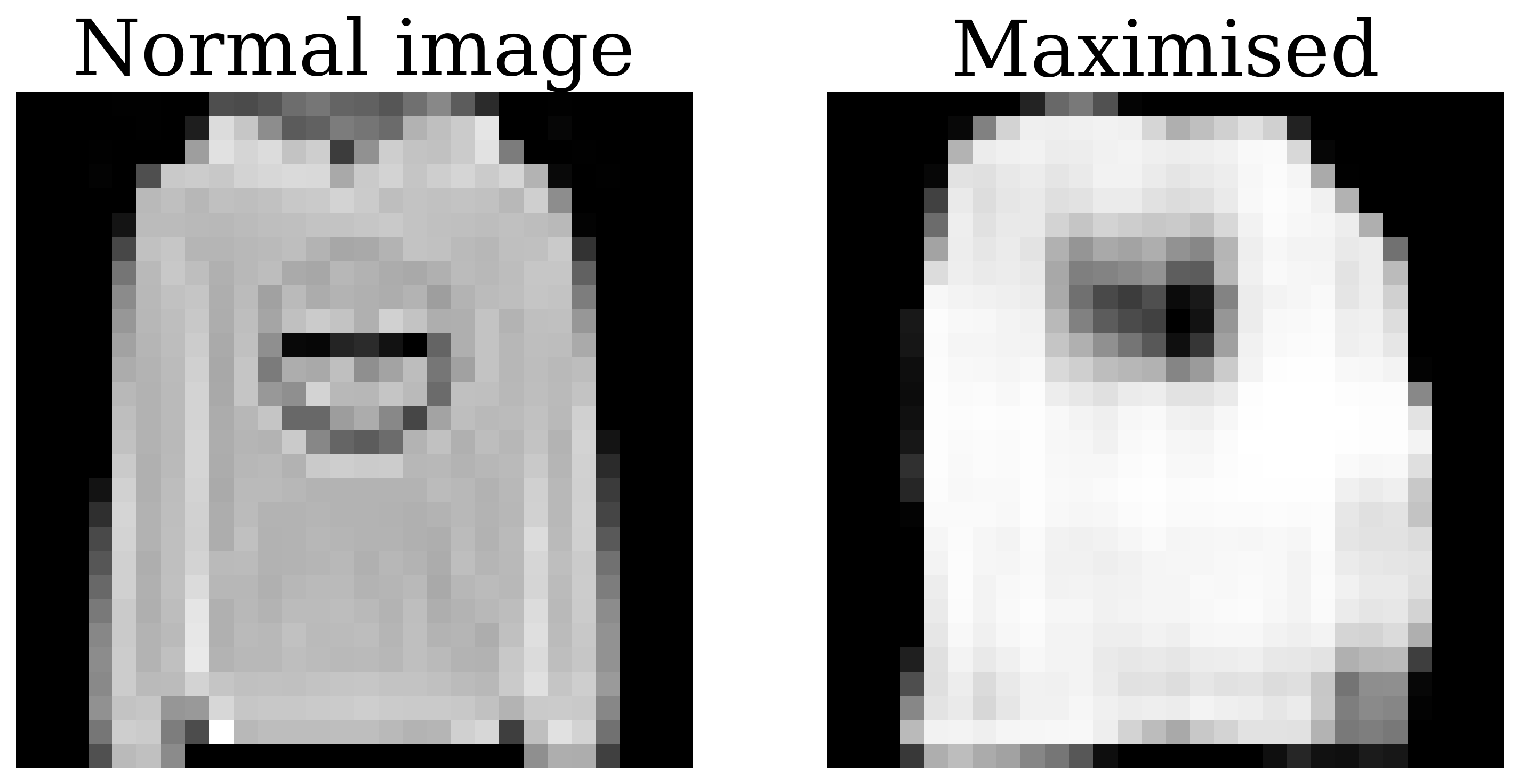}
        \caption{}
        \label{fig:fashion:1}
    \end{subfigure}
    \\[3ex]
    \begin{subfigure}{0.3\textwidth}
        \includegraphics[width=\textwidth]{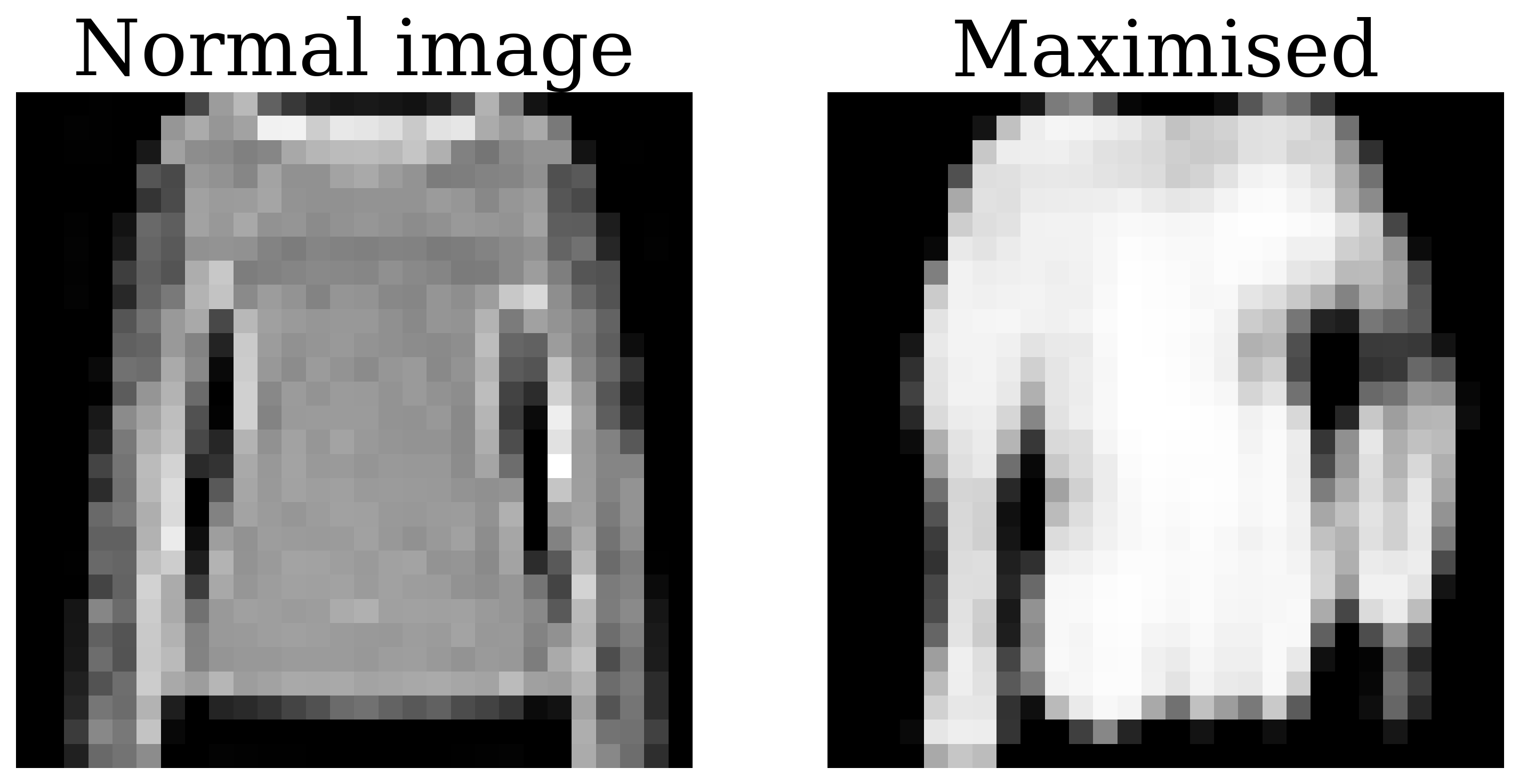}
        \caption{}
        \label{fig:fashion:2}
    \end{subfigure}
    \hspace{1.5cm}
    \begin{subfigure}{0.3\textwidth}
        \includegraphics[width=\textwidth]{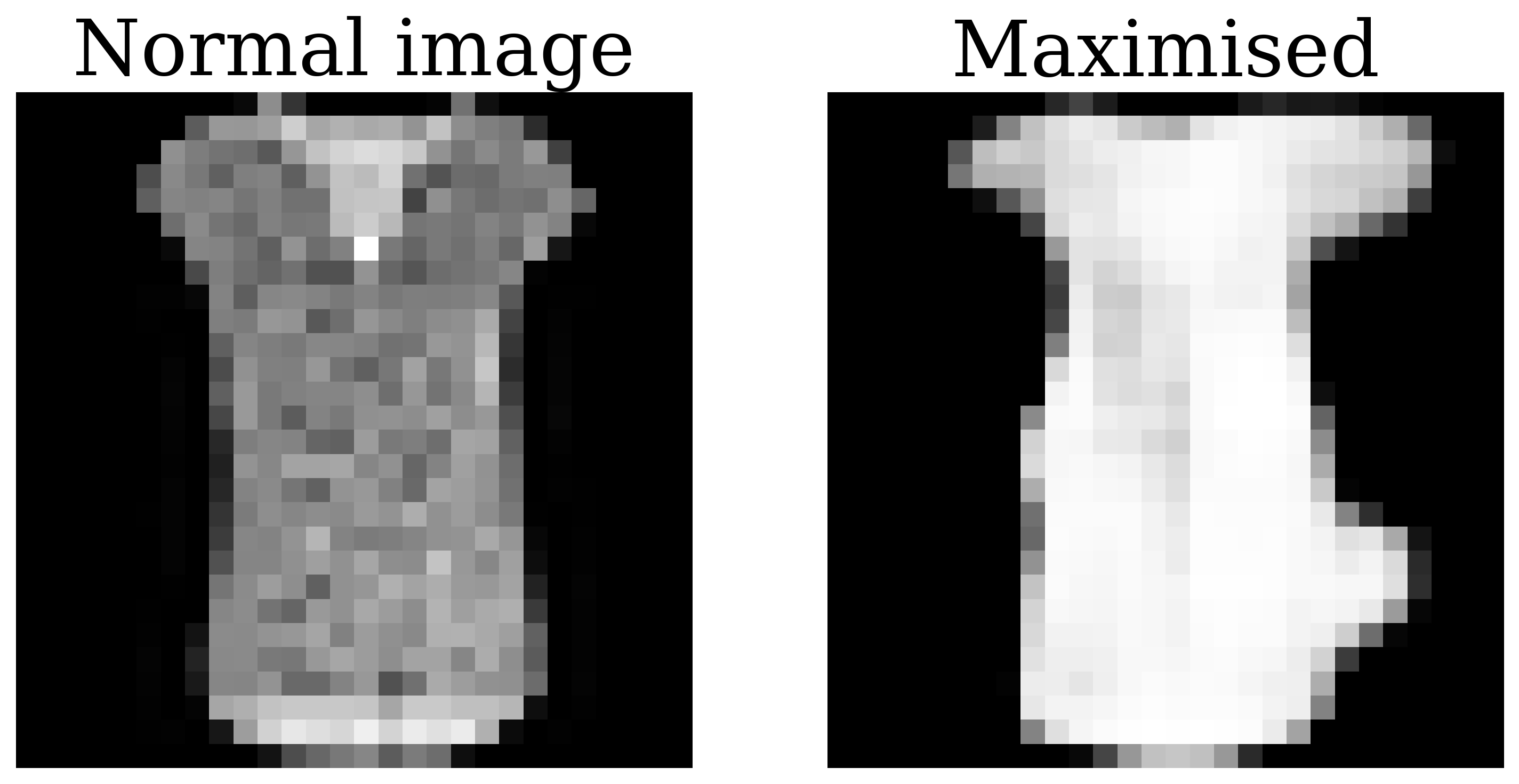}
        \caption{}
        \label{fig:fashion:3}
    \end{subfigure}
    \caption{\label{fig:fashion}Samples from the Fashion-MNIST dataset, with corresponding maximisations.}
\end{figure}

The results for the method described in Sec.~\ref{section:tabular-data} for the MNIST-autoencoder are shown in Fig.~\ref{fig:mnist}, and the results for the Fashion-MNIST classifier are shown in Fig.~\ref{fig:fashion}. We observe that all samples achieve successful maximisation. However, it is also worth noting that most maximised MNIST-samples are often visually very different from their original images.

\subsection{Chess}
\begin{figure}
    \centering
    \begin{subfigure}{0.40\textwidth}
        \includegraphics[width=\textwidth]{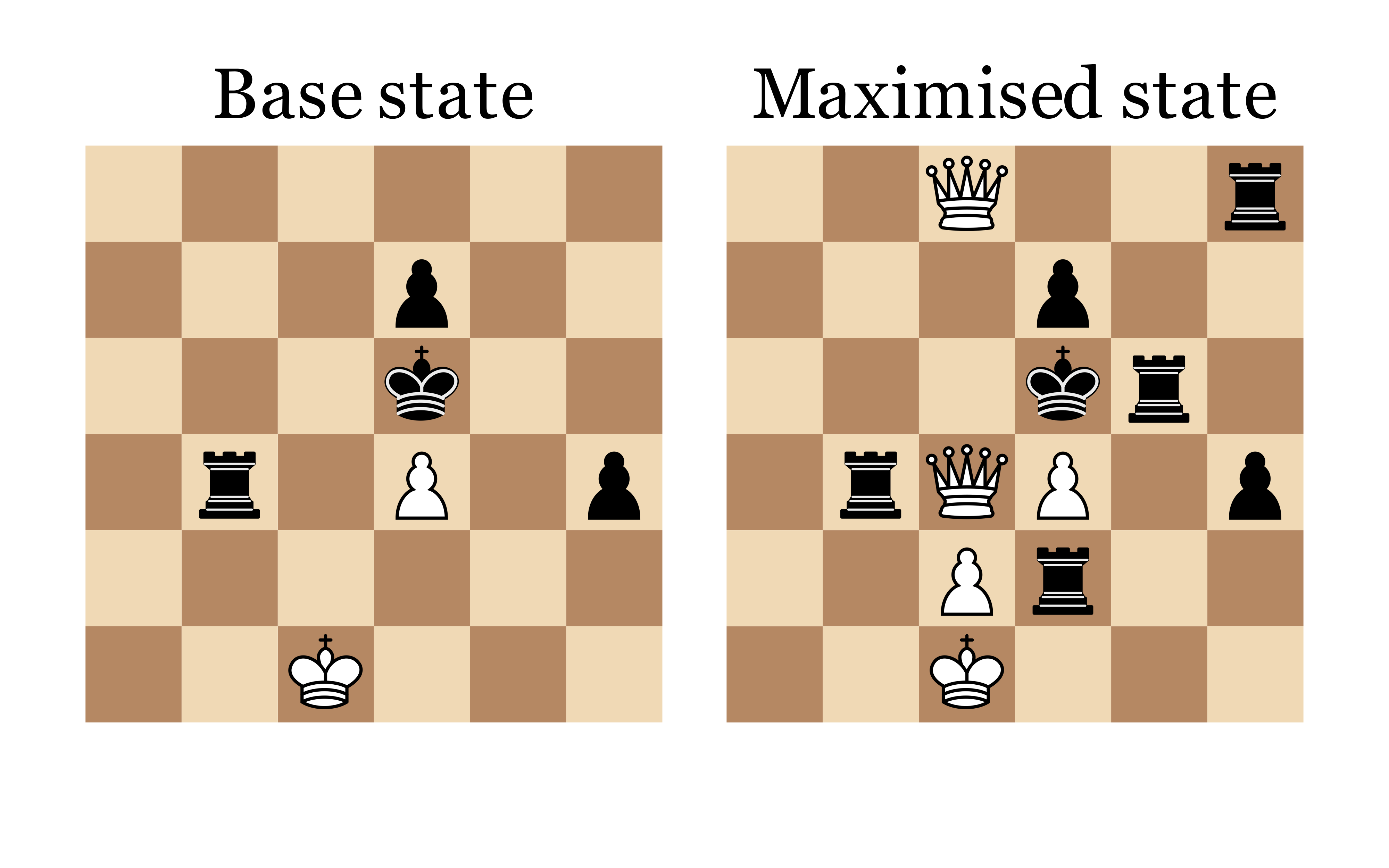}
        \caption{}
        \label{fig:chess:0}
    \end{subfigure}
    \begin{subfigure}{0.40\textwidth}
        \includegraphics[width=\textwidth]{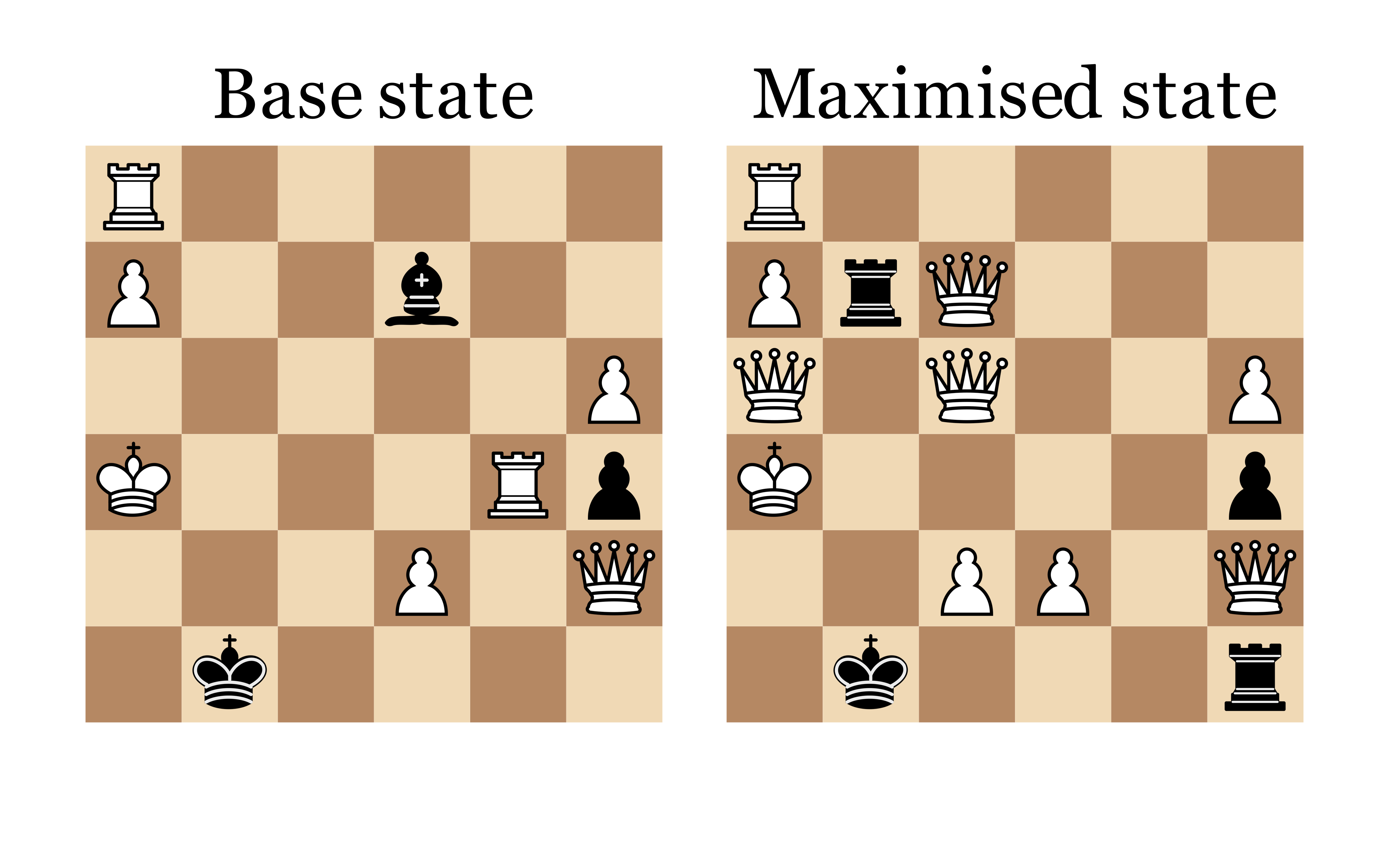}
        \caption{}
        \label{fig:chess:1}
    \end{subfigure}
    \begin{subfigure}{0.40\textwidth}
        \includegraphics[width=\textwidth]{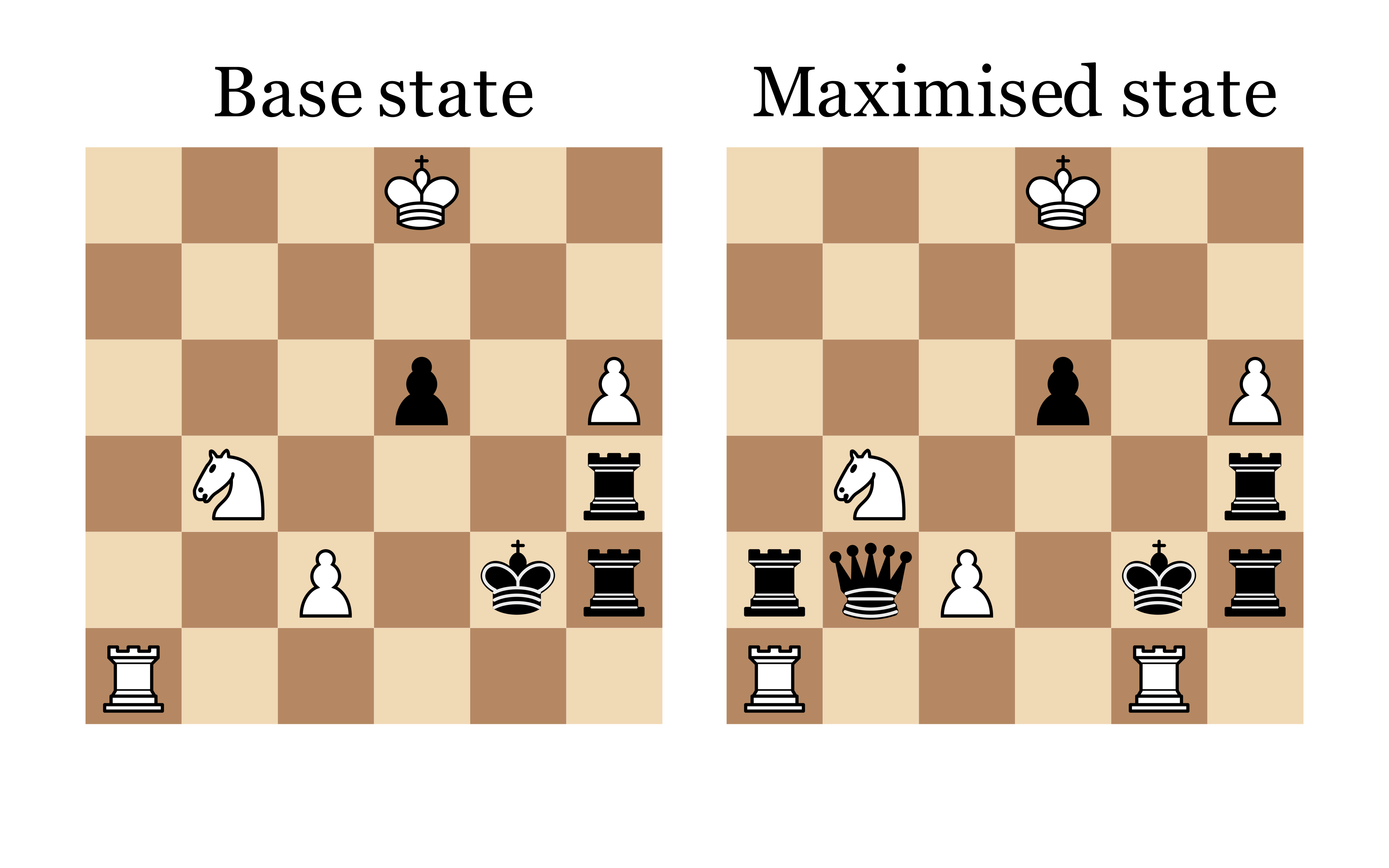}
        \caption{}
        \label{fig:chess:2}
    \end{subfigure}
    \begin{subfigure}{0.40\textwidth}
        \includegraphics[width=\textwidth]{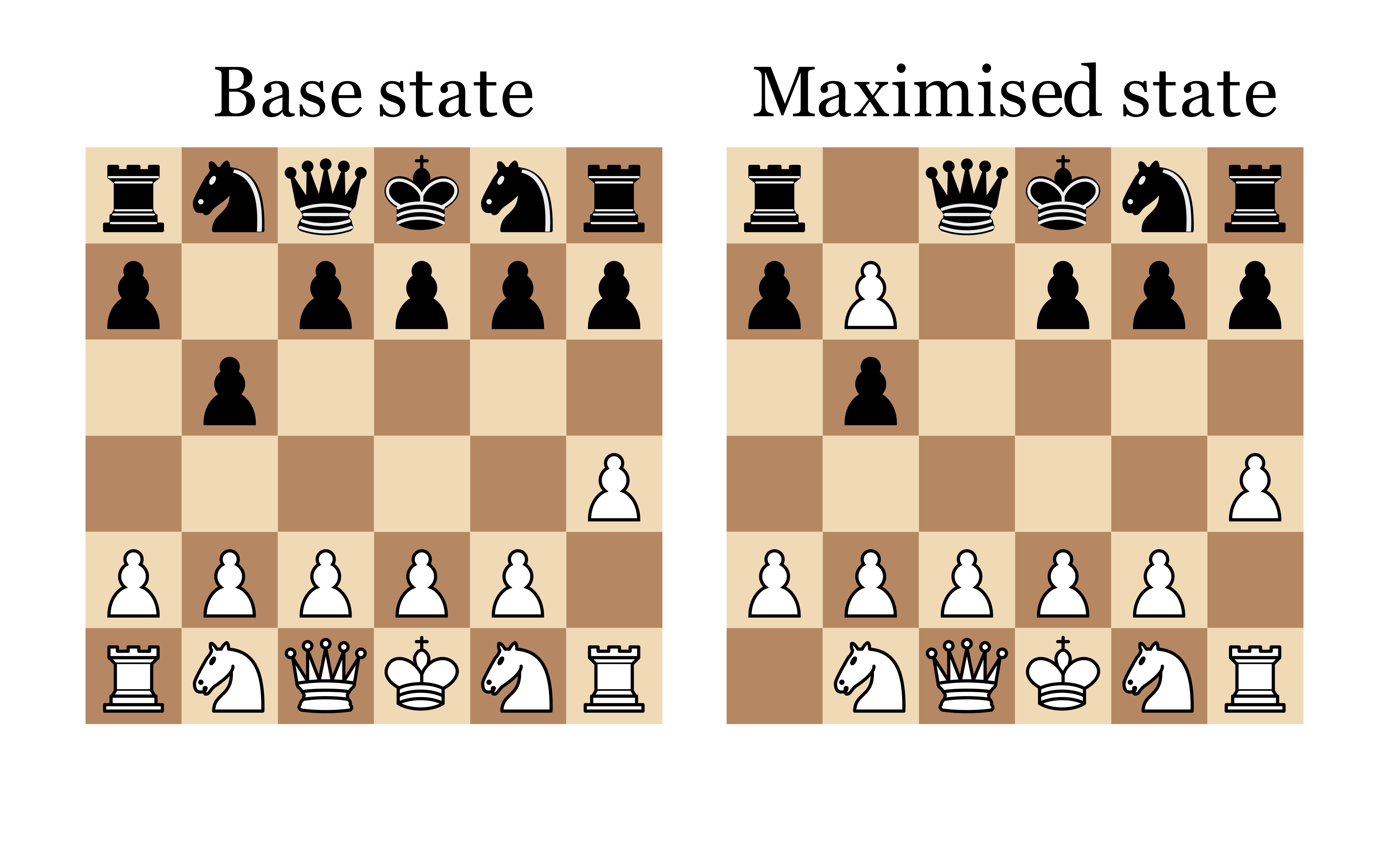}
        \caption{}
        \label{fig:chess:3}
    \end{subfigure}
    \caption{\label{fig:chess}Samples obtained by using the models described in \cite{masterpiece}, with corresponding states where the threat on the opposite's player queen(s) is maximised.}
\end{figure}

The results for the method described in Sec.~\ref{section:chess} for the neural network model trained on 6x6-chess are shown in Fig.~\ref{fig:chess}. We observe that most samples achieve successful maximisation, but that many of these additionally introduce pieces that are not seemingly relevant to the given concept. This is discussed in Sec.~\ref{section:discussion}.

\section{Discussion}\label{section:discussion}
\begin{figure}
    \centering
    \begin{subfigure}{0.18\textwidth}
        \includegraphics[width=\textwidth]{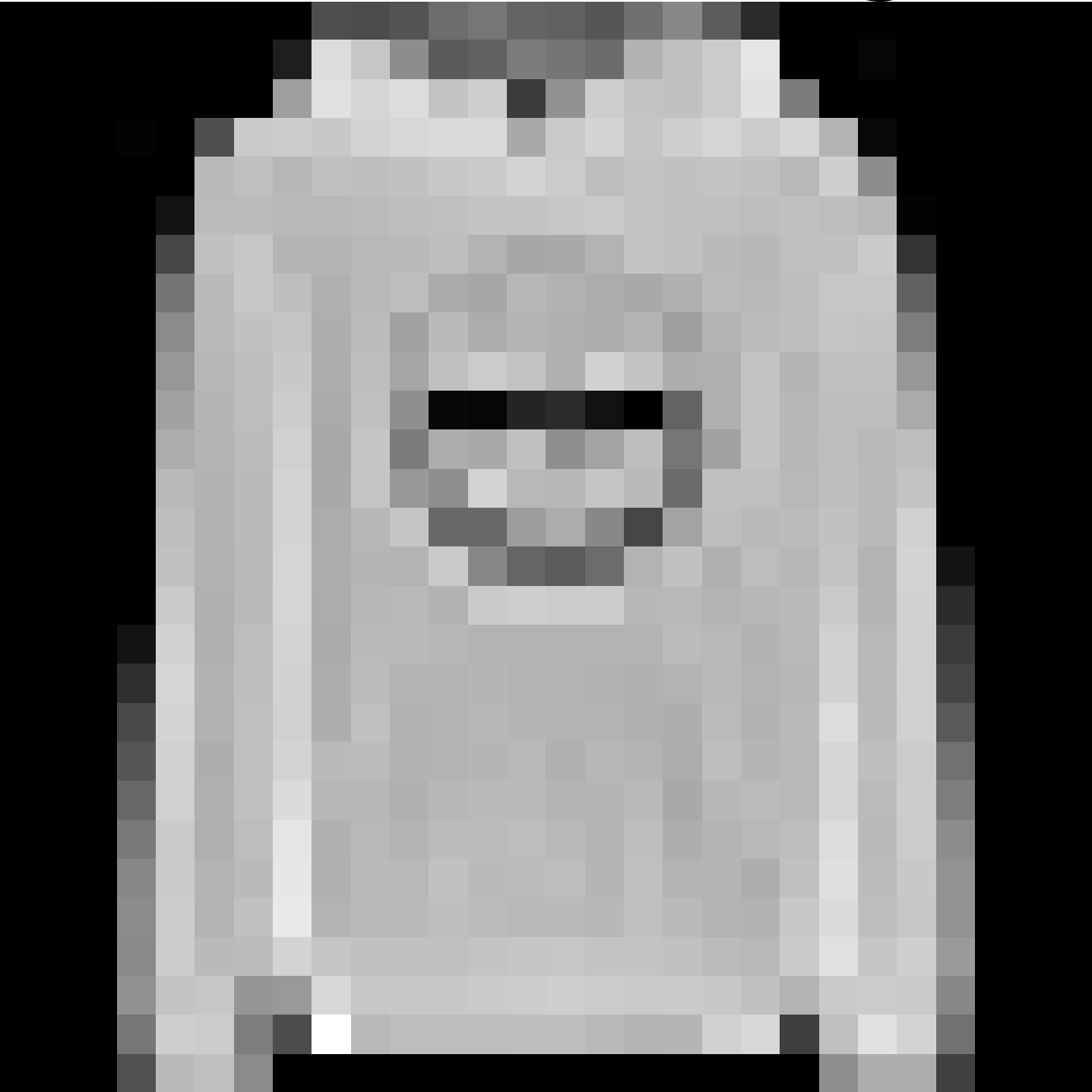}
        \caption{}
        \label{fig:fashion-weights:org}
    \end{subfigure}
    \begin{subfigure}{0.18\textwidth}
        \includegraphics[width=\textwidth]{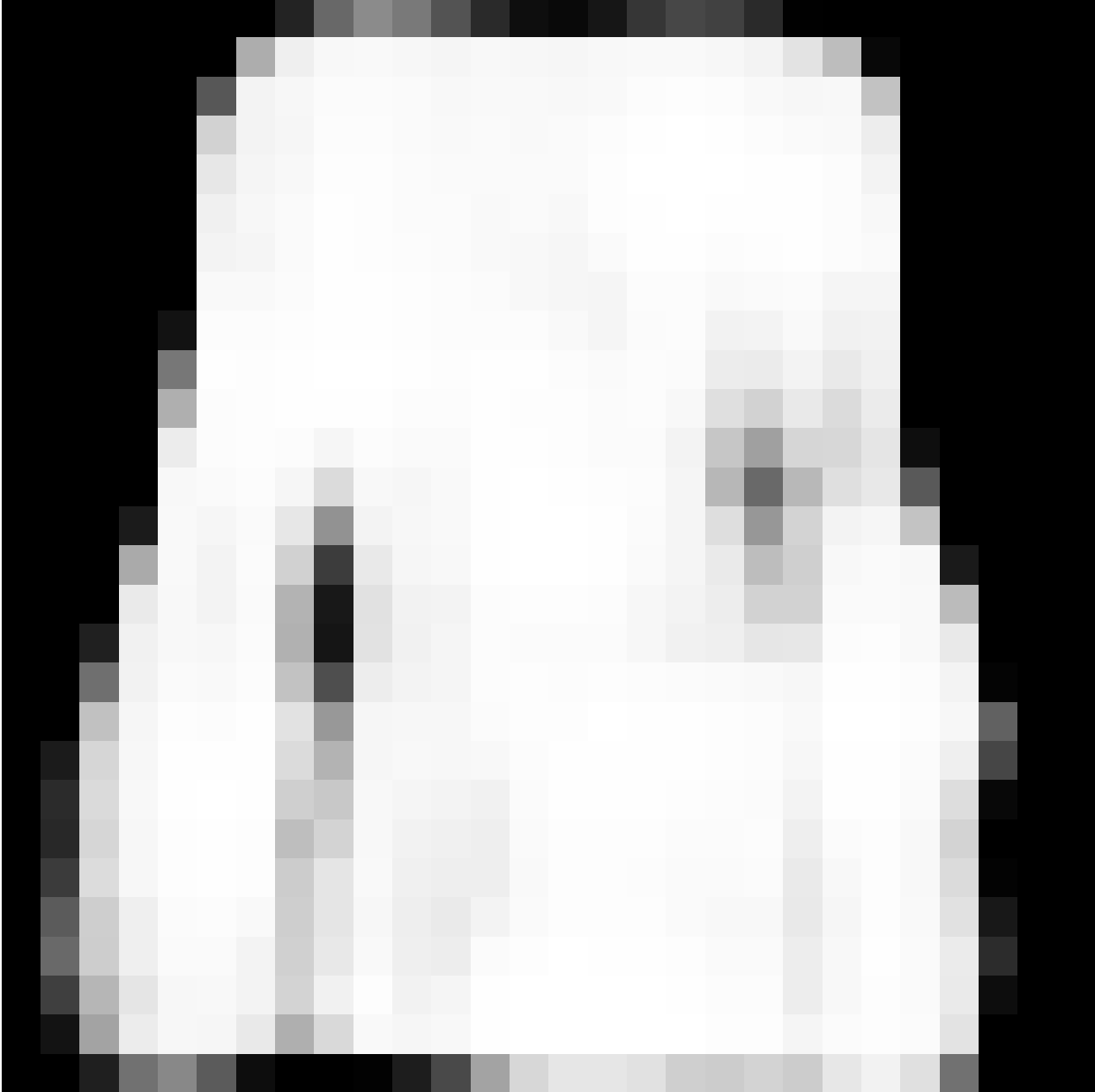}
        \caption{}
        \label{fig:fashion-weights:0}
    \end{subfigure}
    \begin{subfigure}{0.18\textwidth}
        \includegraphics[width=\textwidth]{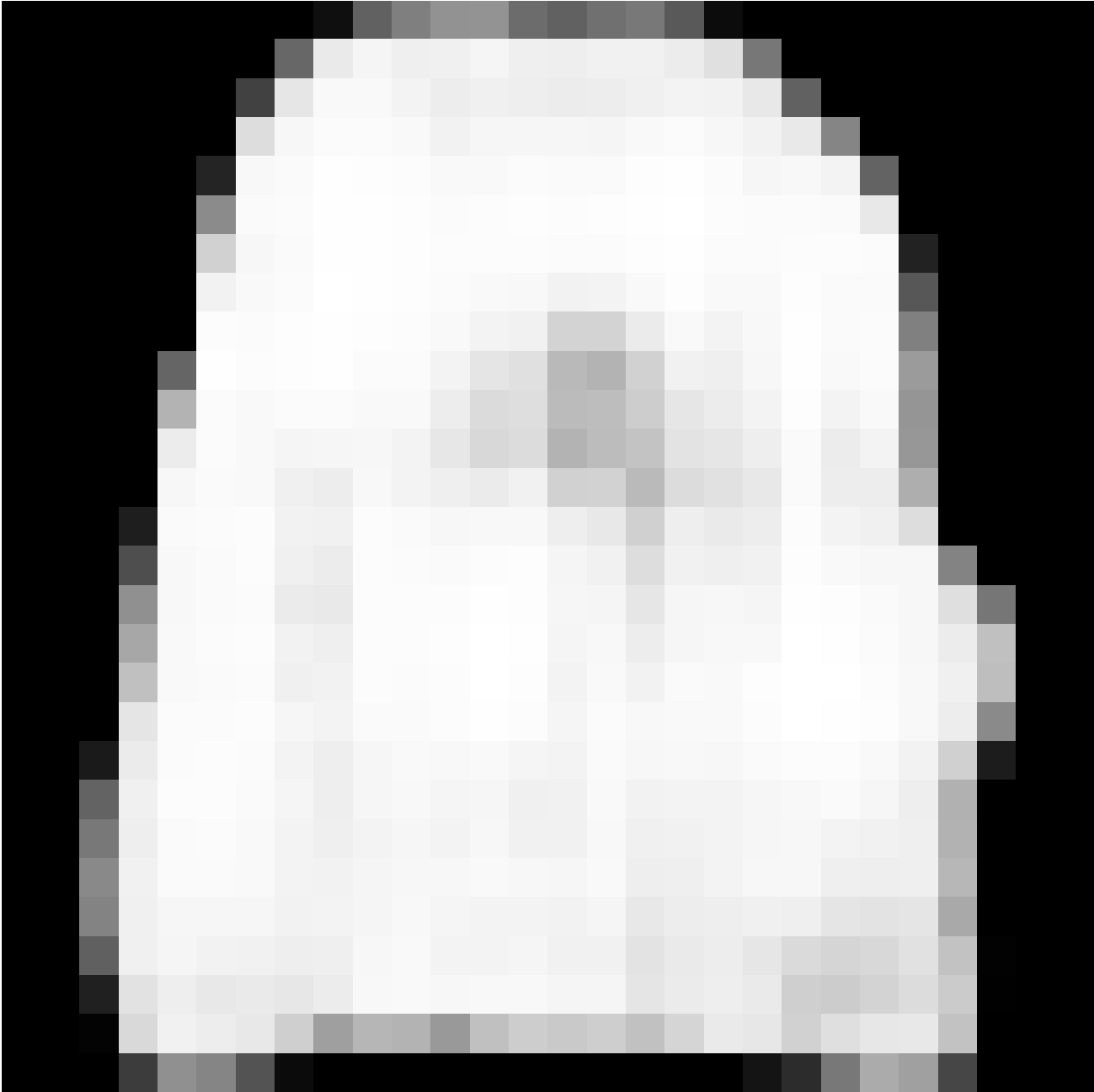}
        \caption{}
        \label{fig:fashion-weights:1}
    \end{subfigure}
    \begin{subfigure}{0.18\textwidth}
        \includegraphics[width=\textwidth]{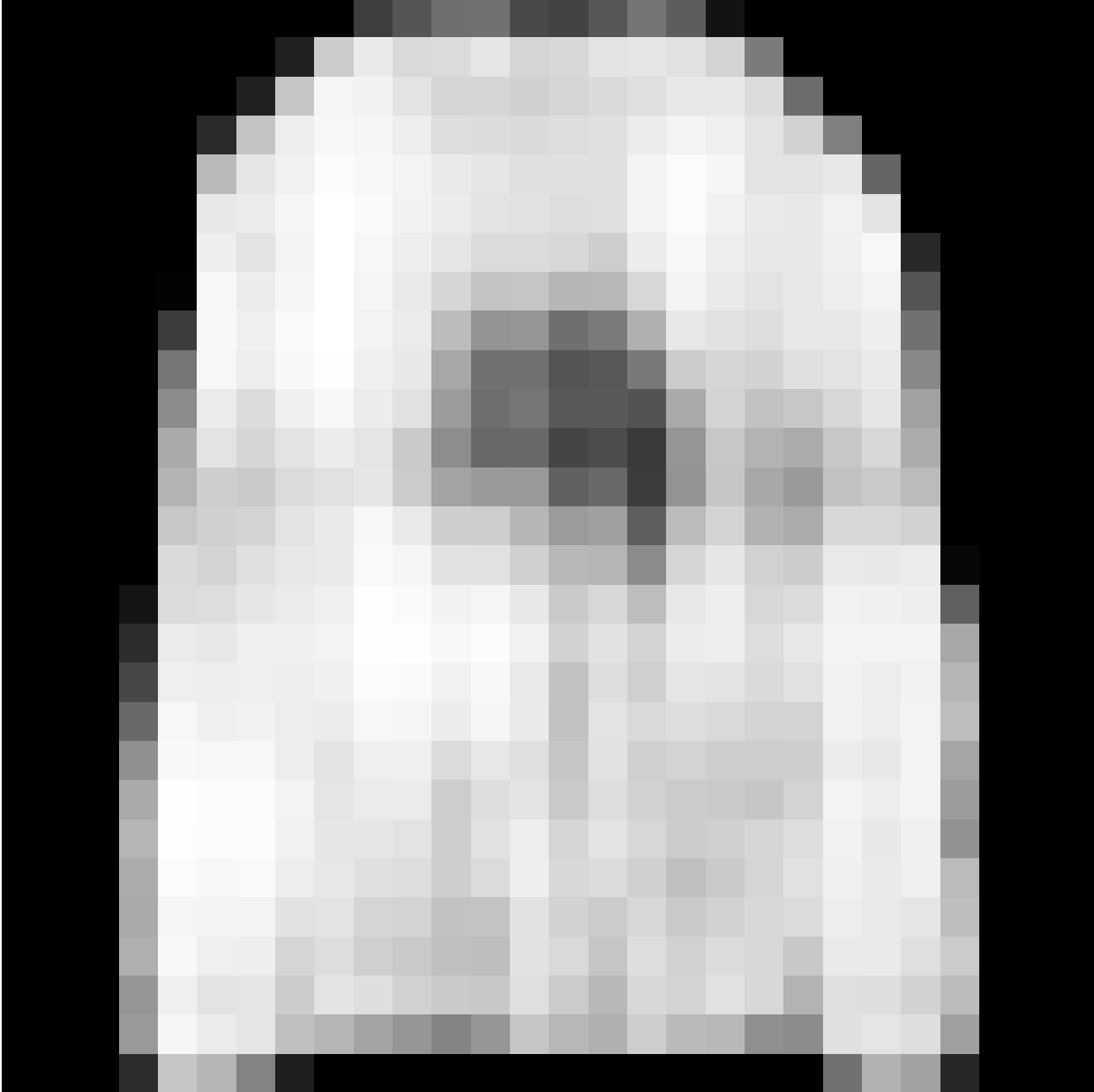}
        \caption{}
        \label{fig:fashion-weights:2}
    \end{subfigure}
    
    \hspace{1.5cm}
    \caption{Maximised samples from the Fashion-MNIST dataset, with (\protect{\subref{fig:fashion-weights:org}}) being the original image, (\protect{\subref{fig:fashion-weights:0}}) being a resulting maximisation with $\lambda_1 = 1.0$, $\lambda_2 = 0.1$, (\protect{\subref{fig:fashion-weights:1}}) $\lambda_1 = 1.0$, $\lambda_2 = 0.75$, and (\protect{\subref{fig:fashion-weights:2}}) $\lambda_1 = 1.0$, $\lambda_2 = 5.0$.}
    \label{fig:fashion-weights}
\end{figure}
We have demonstrated that the proposed method allows for direct visualisation of learned concepts in neural network models over a wide variety of domains, meaning that it presents \emph{how} these models learn to internalise the given concepts. This method is therefore applicable for most cases where one wishes to utilise concept detection to probe for learned knowledge in trained neural network models.

Through the results shown in Table \ref{table:results:housing-dataset}, it is observed that the presented method is suitable for uncovering how entangled features can affect how a given model internalise concepts. When the model is tasked with maximising the ratio of the average amount of bedrooms per person with regard to its intermediate activation space, it also increases the average amount of rooms per person. While this is a logical degree of entanglement, it also means that a ``standard" procedure of concept detection might incorrectly lead one to assume that this ratio is internalised independently of this confounding factor. In this case, the results by the presented methods might suggest that it is more apt to consider these three variables together, even if a valid concept detection result is obtained. These results also indicate that the utility of the proposed method is likely to be significantly higher when applied to models of high complexity, such as multi-layer neural networks. This is because such models often learn complex mappings with several entangled relationships between features, which in turn makes it possible for these relationships to be highlighted by using the proposed method. For simpler regressors, however, these relationships would most likely be trivially available through direct inspection of the learned model itself.

The results shown in Figs.~\ref{fig:mnist} and \ref{fig:fashion} also show that it is possible to generate perturbations to maximise concepts by first mapping a given model's input space to an embedding space. This is useful for models that operate in a input space that is hard to work with directly. This in turn makes it possible to operate in an embedding space that is easier to operate, while still retaining the interpretability and visual capabilities of input spaces such as images.

The results from the chess-playing model, as shown in Fig.~\ref{fig:chess}, highlight that it is possible to create valid maximisations adhering to multiple strict constraints. While most samples present a successful maximisation of the given concept (as shown in Figs.~\ref{fig:chess:0}, \ref{fig:chess:1}, \ref{fig:chess:3}), we also observe that it does not succeed in doing so in some cases, as exemplified through Fig.~\ref{fig:chess:2}. Empirically, this was observed to occur more frequently with the method to the chess model as opposed to the other modalities. We hypothesise that this is due to the difficulty of finding a valid perturbation that successfully maximises the given concept, while ensuring that the resulting perturbation abides by all rules described in Sec.~\ref{section:chess}. This is also implied by the fact that most of the samples presented in Fig.~\ref{fig:chess} also introduce a notable amount of pieces that are not relevant for the position at hand.\footnote{See e.g. Fig.~\ref{fig:chess:3}, where White's leftmost rook and Black's leftmost knight are removed. The position otherwise provides a valid maximisation.} While this might be attributable to some aspect of the model's learned representation of the concept, it is in this case strongly hypothesised to be due to the discreteness of chess as an input space.

While the method is very generalisable, it can in many cases be difficult to find the right balance between highlighting a given concept, and preserving the original structure of the input sample itself. In practice, this amounts to find an adequate tuning of $\lambda_1$ and $\lambda_2$. An example of this can be seen in Fig.~\ref{fig:fashion-weights}. Here, the method produces wildly different maximisations for the different weightings of the minimisation objective described in Eq.~\ref{equation:generic-backprop}. While this is a problem in some cases, it also shows that the method facilitates the generation of various samples that maximise the relevant concepts for almost all constraints. Additionally, since this method is relatively inexpensive to perform, it is also possible to generate many such perturbations with different weights, in order to consider a larger variety of samples.

% Looking at how different elements of input space are connected. ``Forcing" model to maximise some concept might also cause it to change some elements that are at not directly related to the concept, but might be related to the input features that in turn are related to the concept. (ex. amount of rooms being changed when wanting to maximise the amount of bedrooms per person) Tells us that model probably doesn't consider AveBedrooms/Occupancy directly, but includes AveRooms as well, which in turn is due to AveRooms and AveBedrooms being highly correlated. This in turn gives a clearer picture as to what is included in the model's internal representation of the concept. 

\section{Conclusion}
We have presented a method that allows for visualisations of learned concepts through concept maximisation. This is relevant for obtaining a deeper understanding of how a given neural network model learns to internalise important concepts, and for easily presenting these internalised representations in the model's own input space. The method is generalisable to most domains, allowing for easy applicability for a diverse set of problems independently of both network architecture and input structure. Finally, the method draws from the strengths of concept detection, which means that it can be applied to most pre-trained models, without requiring expensive training procedures to be performed. In this vein, an interesting future work for this method would be to apply it to problems of higher complexity, such as state of the art image classifying models, or large language models.

\bibliographystyle{IEEEtran}
\bibliography{IEEEabrv, refs}

\end{document}